\documentclass[11pt]{article}

\usepackage[preprint]{acl}

\usepackage{times}
\usepackage{latexsym}

\usepackage[T1]{fontenc}

\usepackage[utf8]{inputenc}

\usepackage{microtype}

\usepackage{inconsolata}

\usepackage{graphicx, multirow, multicol, url, booktabs, amsmath, bm, cleveref, subcaption, enumitem, array, comment}

\crefformat{footnote}{#2\footnotemark[#1]#3}

\title{Towards Self-Referential Analytic Assessment: A Profile-Based Approach to L2 Writing Evaluation with LLMs}

 \author{Stefano Bannò, Kate Knill, Mark Gales \\
         ALTA Institute, Department of Engineering, University of Cambridge (UK) \\ \texttt{\{sb2549,kmk1001,mjfg100\}@cam.ac.uk}}

\begin{document}
\maketitle
\begin{abstract}
Automated essay scoring (AES) research often relies on rank-based correlation metrics to validate analytic assessment. However, such metrics obscure both intrinsic intercorrelations among analytic dimensions that arise from the structure of writing proficiency itself and halo effects, whereby holistic impressions bleed into fine-grained component scores. As a result, high correlations may mask a system’s true diagnostic behaviour. In this study, we propose a novel self-referential assessment evaluation framework that focuses on identifying intra-learner strengths and weaknesses rather than assessing inter-learner rankings. We conduct experiments on the publicly available ICNALE GRA, a uniquely dense second-language writing dataset annotated holistically and analytically by up to 80 trained raters. To obtain reliable reference scores, we apply two-facet Rasch modelling to calibrate rater severity and derive fair average scores across ten analytic aspects and holistic proficiency. We compare the analytic scoring performance of human operational raters and three large language models (LLMs) in a zero-shot setting. Our results show that LLMs tend to outperform single human raters in identifying relative weaknesses (negative feedback) across several proficiency aspects, while human raters remain stronger at identifying relative strengths (positive feedback). Overall, our findings highlight the limitations of rank-based evaluation for analytic assessment and demonstrate the value of intra-learner, profile-based methods for assessing and deploying LLMs in AES.
\end{abstract}

\section{Introduction}

Automated essay scoring (AES) has become a cornerstone of second language (L2) proficiency assessment, utilising computational systems to evaluate learner writing in educational contexts.~\cite{klebanov2022automated}. AES systems have primarily targeted holistic assessment, focusing on assigning a single score that reflects a learner's overall writing proficiency. However, driven by the need for more actionable pedagogical insights, the field has recently expanded its focus to analytic scoring, i.e., the evaluation of distinct aspects of proficiency~\cite{li2024essay}. The advent of large language models (LLMs) has revolutionised this landscape. Recent work has successfully exploited LLMs for both holistic~\cite{mizumoto2023gpt} and analytic scoring~\cite{banno-etal-2024-gptfull}. This shift offers the potential for scalable, multi-dimensional assessment that was previously difficult to achieve with traditional supervised models.

\begin{figure*}[ht!]
    \centering
\includegraphics[width=0.6\textwidth]{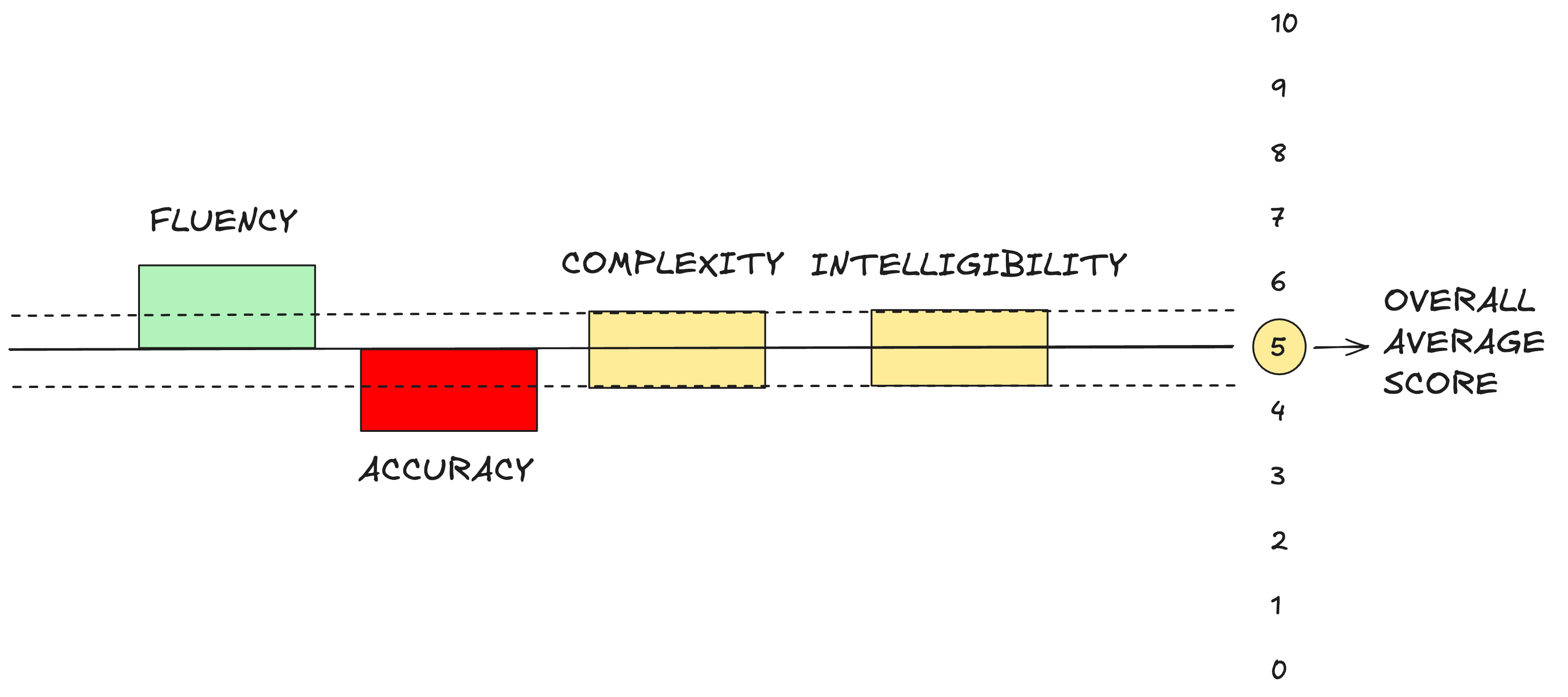}
    \caption{Illustration of proposed self-referential analytic assessment approach.}
    \label{fig:ipsative_diagram}
\end{figure*}

Despite these advances, current evaluation practices predominantly rely on rank-based metrics, such as Pearson’s correlation coefficient (PCC), Spearman’s rank coefficient (SRC), or agreement metrics, such as Quadratic Weighted Kappa (QWK)~\cite{yannakoudakis-cummins-2015-evaluating}. While these metrics effectively measure a system's ability to rank learners globally (i.e., inter-learner agreement), they may not be suitable to validate multi-dimensional analytic assessment. In addition to intrinsic intercorrelations among analytic dimensions, a major interfering factor is the halo effect~\cite{thorndike1920constant, engelhard1994examining}, whereby a learner's general proficiency influences judgments across specific sub-scales. As we show in this study, an LLM can achieve high correlations on analytic aspects simply by acting as a proxy for holistic proficiency, without accurately detecting the nuances that distinguish different analytic proficiency dimensions. Consequently, high rank agreement may not guarantee diagnostic validity. To address this limitation, we propose shifting the analytic evaluation paradigm from \emph{normative} (i.e., ranking learners against each other) to \emph{self-referential} assessment, in which a learner’s analytic skills are evaluated within the same learner profile.
Self-referential analytic assessment focuses on intra-learner variability, identifying aspects where a student performs significantly better or worse than their average proficiency (see Figure \ref{fig:ipsative_diagram}), thus offering a stricter and more pedagogically relevant test of an LLM's analytic assessment capabilities.

Ensuring the interpretability of such an evaluation requires a reference signal that is internally consistent, psychometrically calibrated, and representative of expert judgement under the known interdependence of analytic dimensions. Standard datasets often suffer from rater subjectivity and noise, which can obscure the genuine signal. We mitigate this by leveraging the publicly available ICNALE GRA~\cite{ishikawa2024icnale} (see Section \ref{sec:data}), an L2 learner dataset annotated by 80 raters holistically and analytically. We employ Rasch modelling~\cite{linacre1989many} to rigorously filter raters based on infit statistics, constructing a fair average reference score derived solely from highly consistent human raters. 

In this paper, we evaluate three LLMs in a zero-shot setting alongside operational human raters using this self-referential framework. Our contributions are as follows:

\begin{enumerate}

\item We demonstrate the limitations of traditional rank-based metrics in analytic assessment, showing that high correlations often mask a lack of diagnostic precision due to both intrinsic intercorrelations and the halo effect.

\item We introduce a novel self-referential analytic assessment evaluation method that disentangles absolute proficiency from relative profile deviations, creating a dual classification task for positive and negative feedback.

\item We report experimental results showing that, within this self-referential framework, LLMs outperform a single operational rater in identifying relative weaknesses (negative feedback), while human raters retain an advantage in identifying relative strengths (positive feedback).\end{enumerate}

\section{Related work}



Traditionally, AES systems have mainly 
focused on overall holistic assessment. This holistic orientation has characterised AES since its inception in the 1960s~\cite{page1966imminence} and has been predominant up to today in recent studies investigating L2 assessment with LLMs~\cite{mizumoto2023gpt, yancey-etal-2023-ratingfull}. 

Research on automated analytic assessment and feedback emerged later, with only a few early exceptions~\cite{page1997computer, shermis2002trait}. The advent of neural approaches led to a growing body of work targeting specific traits of written production, such as organisation, content, word choice, sentence fluency, and narrativity~\cite{hussein2020trait-based, mathias-bhattacharyya-2020-neural, ridley2021, kumar-etal-2022-many, do-etal-2023-prompt}. As with holistic assessment, LLMs have also been successfully applied to analytic assessment. For instance, \citet{naismith-etal-2023-automatedfull} investigated the use of GPT-4 to predict coherence-related scores. \citet{banno-etal-2024-gptfull} explored zero-shot analytic assessment of L2 writing proficiency using GPT-4 across CEFR-aligned proficiency aspects~\cite{cefr2020}. \citet{sessler2025can} examined multiple open- and closed-source LLMs for analytic scoring of German student essays across ten traits, including content-related, syntactic, and formal dimensions. \citet{wang-etal-2025-llms-perform} investigated LLMs’ ability to conduct multi-dimensional analytic writing assessment by assigning scores and generating feedback comments across nine analytic aspects, finding that LLMs can produce generally reliable analytic assessments. In another recent work, \citet{yoo-etal-2025-dress} introduced a new dataset annotated across three analytic traits, i.e., content, language, and organisation, combining LLM-based scoring with synthetic data generation. Using the ICNALE GRA essays, \citet{yamashita2024application} investigated the use of GPT-4 for analytic assessment along complexity, accuracy, and fluency dimensions. Their study primarily adopts a many-facet Rasch measurement framework to analyse rater severity, consistency, and potential biases when treating GPT-4 as an additional rater alongside humans.

Automated analytic scoring has largely inherited the evaluation metrics used for holistic assessment. The studies reviewed above typically rely on rank-based metrics, such as PCC or SRC, as well as agreement measures such as QWK~\cite{yannakoudakis-cummins-2015-evaluating}, under a normative assessment framework, whereby learners are ranked against each other for each analytic proficiency aspect. In our work, we propose shifting the evaluation paradigm for analytic assessment from normative evaluation to self-referential interpretation of analytic profiles, in which analytic traits are evaluated within an individual learner's performance profile. In the psychometric literature, a similar approach has been theorised as \emph{ipsative} assessment~\cite{cattell1944psychological, clemans1956analytical}. However, in some applied assessment traditions, the same term instead denotes comparison with an individual’s previous performances~\cite{hughes2011towards}, rather than within-individual, self-referenced comparison across analytic dimensions. For this reason, we prefer the term \emph{self-referential}. 

Self-referential analytic assessment directly supports diagnostic feedback by identifying relative strengths and weaknesses within a learner, rather than across a population. To the best of our knowledge, this is the first study to introduce a self-referential evaluation framework for analytic assessment of L2 proficiency and to implement automated systems for this purpose. The closest conceptual precedent is \citet{berninger2010listening}, who employed within-learner comparisons across language modalities (listening, speaking, reading, and writing), operationalising relative strengths and weaknesses as deviations from a learner’s own mean performance. While their approach focused on modality-level  L1 language skills in a psychometric testing context and did not use any automatic systems, the present study applies a self-referential perspective within AES of L2 writing by comparing learners’ performance across fine-grained analytic proficiency aspects.

\section{Experimental setup}

In this section, we first describe the dataset used in our experiments and outline the procedures for rater selection and score calibration. We then present details on LLM prompting and the method for extracting scores from the models. Next, we discuss the limitations of rank-agreement metrics when evaluating analytic assessment. Finally, we introduce our self-referential analytic assessment system.

\subsection{Data}
\label{sec:data}

The International Corpus Network of Asian Learners of English (ICNALE) Global Rating Archive (GRA)~\cite{ishikawa2020, ishikawa2024icnale} is a publicly available L2 learner dataset derived from a subset of ICNALE~\cite{ishikawa2011}, containing learner essays and speeches, of which we use only the essay section for our experiments. A notable feature of this dataset is its English as a Lingua Franca (ELF) assessment perspective~\cite{seidlhofer2005elf}, which evaluates English as a communicative tool among non-native speakers instead of relying on native-speaker norms. Although the dataset is relatively small ($N=140$), each essay has been annotated by 80 trained raters representing diverse first‑language (L1) and occupational backgrounds.\footnote{Four essays were written by L1 speakers; we did not discard them from the dataset.}

In addition to holistic scores, the essays have been annotated across three macro-aspects, which in turn comprise ten analytic rating aspects: 

\begin{itemize}
    \item \textbf{Language} (\emph{Intelligibility}, \emph{Complexity}, \emph{Accuracy}, and \emph{Fluency});
    \item \textbf{Content} (\emph{Comprehensibility}, \emph{Logicality}, \emph{Sophistication}, and \emph{Purposefulness});
    \item \textbf{Attitude} (\emph{Willingness to communicate} and \emph{Involvement}).
\end{itemize}
 
 The analytic scores range from 0 to 10, whereas the holistic scores range from 0 to 100; both scales allow midpoint values.\footnote{We identified four anomalies in the original scores: for \emph{Intelligibility}, a score of 1.2 (recoded as 1.0) and a score of 6.6 (recoded as 6.5); for \emph{Complexity}, a score of 19 (recoded as 10); and for \emph{Involvement}, a score of 8.6 (recoded as 8.5).} Human raters were instructed to evaluate the essays holistically first, followed by analytic scoring.

To the best of our knowledge, this makes ICNALE GRA the only publicly available L2 learner writing dataset that supports a fully crossed rater-essay design with more than two raters, comprising 123,195 ratings.\footnote{\label{fn:missing}That is, $(140\ \text{essays} \times 80\ \text{raters} \times 11\ \text{proficiency aspects}$, i.e., ten analytic and one holistic$) - 5$ missing ratings. Five rating data points are missing from the original scores of the 80 raters; however, there are no missing values in the subset of 12 selected raters (see Section~\ref{sec:data}) used in our experiments.}

\subsubsection{Raters selection}
\label{sec:rater_selection}

In a previous study by the curator of the dataset, focusing specifically on the ICNALE ratings \cite{ishikawa2023effects}, inter-rater reliability is evaluated using Cronbach’s $\alpha$~\cite{cronbach1951coefficient}. This statistic primarily reflects relative consistency, capturing the extent to which raters produce similar rank orderings of test-takers, while being insensitive to systematic differences in absolute scoring levels. As such, Cronbach’s $\alpha$ is a measure of internal consistency rather than a direct measure of inter-rater reliability~\cite{krippendorff2004reliability, lombard2006content}; it reflects the extent to which raters are measuring the same underlying construct.

For this reason, we instead adopt Krippendorff's $\alpha$~\cite{krippendorff2011computing}, which is more commonly used for inter-rater reliability, as it explicitly quantifies absolute agreement among raters, i.e., the degree to which raters assign identical scores to the same test-taker. Moreover, this metric is well suited to ordinal data, multiple raters, and missing values (see Note \ref{fn:missing}). Table \ref{tab:reliability_k} reports inter-rater reliability in terms of Krippendorff's $\alpha$: column \emph{All} shows the estimate computed over all 80 raters, while column \emph{Selected} reports the estimate on 12 raters, after applying the rater selection procedure described below.
\begin{table}[ht!]
\small
    \centering
    \begin{tabular}{l|l|cc}
        \hline
        \textbf{Macro-} & \textbf{Aspect} 
        & \multicolumn{2}{c}{\bm{$\alpha$}} \\
        \cline{2-4}
        \textbf{aspect} & & \textbf{All} & \textbf{Selected} \\
        \hline
        \multirow{4}{*}{\textbf{Language}} 
        & Intelligibility & 0.270 & \textbf{0.531} \\
        & Complexity     & 0.252 & \textbf{0.542} \\
        & Accuracy       & 0.278 & \textbf{0.505} \\
        & Fluency        & 0.256 & \textbf{0.526} \\
        \hline
        \multirow{4}{*}{\textbf{Content}} 
        & Comprehensibility & 0.257 & \textbf{0.538} \\
        & Logicality        & 0.237 & \textbf{0.499} \\
        & Sophistication    & 0.225 & \textbf{0.517} \\
        & Purposefulness    & 0.221 & \textbf{0.502} \\
        \hline
        \multirow{2}{*}{\textbf{Attitude}} 
        & Willingness & 0.202 & \textbf{0.465} \\
        & Involvement                & 0.172 & \textbf{0.409} \\
        \hline
        \hline
        & Holistic & 0.297 & \textbf{0.579} \\
        \hline
    \end{tabular}
    \caption{Krippendorff's $\alpha$ reliability estimates on ICNALE GRA.}
    \label{tab:reliability_k}
\end{table}
As can be seen, the low Krippendorff’s $\alpha$ values for \emph{All} indicate substantial disagreement among raters, which may be attributable to differences in severity, inconsistent category use, or divergent interpretations of the rating criteria.

To ensure the reliability of the rating data, we evaluate each of the 80 raters' consistency using the infit statistic derived from a two-facet Rasch model~\cite{linacre1989many}, i.e., one model for each of the 11 analytic and holistic language aspects, with the two facets being rater severity and learner ability. To do this, we use the Many-Facet Rasch Model estimation function, \texttt{tam.mml.mfr}, within \texttt{R}'s \texttt{TAM} (Test Analysis Modules) package.\footnote{\url{cran.r-project.org/web/packages/TAM}} To accommodate midpoints, analytic scores are fed into the model after multiplying them by 2, whereas holistic scores are binned into 11 categories from 0 to 10, as described in \citet[p. 33]{ishikawa2024icnale}.

Infit measures how well a rater's scores match the model's expectations, giving more weight to items that are close to a person's ability level. Mathematically, for rater $r$ and item $i$, the infit mean-square is computed as:

\begin{equation}
\text{Infit}_r = \frac{\sum_i w_{ri} (X_{ri} - E_{ri})^2}{\sum_i w_{ri}}
\end{equation}

where $X_{ri}$ is the observed rating, $E_{ri}$ is the Rasch model-predicted rating, and $w_{ri}$ is a weight reflecting the item's information (i.e., its expected variance). An infit value between 0.5 and 1.5 indicates that a rater's judgments are consistent with the model, whereas values substantially above 1.5 or below 0.5 indicate over- or under-discrimination, respectively~\cite{linacre2002infit}. First, we retained only raters with infit equal to or higher than 0.5 and equal to or less than 1.5, ensuring that all included raters provided reasonably consistent scores across items. Subsequently, we identified the intersection of acceptable raters across all language aspects, resulting in a subset of 12 raters (with 18,480 ratings in total) that satisfied the infit criterion for every aspect considered.

As can be observed in column \emph{Selected} in Table~\ref{tab:reliability_k}, rater selection based on infit statistics leads to a substantial increase in observed reliability. This procedure identifies a subset of raters whose scoring behaviour falls within acceptable Rasch infit bounds and can therefore be used for score calibration and subsequent modelling. From this point onwards, all references to ``raters'' pertain exclusively to the subset of 12 selected raters.

\subsubsection{Score calibration}
\label{sec:score_calib}

To obtain a fair average score $E(X_n)$ for a learner $n$, we fit a two-facet Rasch model for each language aspect, this time only with the 12 selected raters. The net cumulative difficulty barrier, $\Phi_{net, k}$, for achieving a score of $k$ combines the average rater severity ($\bar{\lambda}$) and the global cumulative step difficulties ($\tau_m$):
\begin{equation}
\label{eq:net_difficulty}
    \Phi_{net, k} = \left( \sum_{m=1}^{k} \tau_m \right) + k \cdot \bar{\lambda}
\end{equation}
where $\sum_{m=1}^{k} \tau_m$ is the cumulative step difficulty, defined as 0 for $k=0$; $\tau_m$ is the estimated global step difficulty parameter (the difficulty of moving from score $m-1$ to $m$); $k$: the score category index ($k=0, 1, 2, \dots, K$); $\bar{\lambda}$ is the mean severity of all selected raters.

The probability of learner $n$ achieving a score $k$ is calculated as the exponent of the difference between the learner's ability and the net cumulative difficulty, normalised by the sum of probabilities across all possible categories:
\begin{equation}
\label{eq:probability}
    P(X_n=k) = \frac{\exp \left( k\theta_n - \Phi_{net, k} \right)}{\sum_{c=0}^{K} \exp \left( c\theta_n - \Phi_{net, c} \right)}
\end{equation}
where $\theta_n$ is the estimated ability of learner $n$ and $K$ is the maximum category index.

The final expected score $E(X_n)$ is the sum of the products of each category score $s_k$ and its probability $P(X_n=k)$. Since, for the analytic aspects, the raw scores were multiplied by 2 prior to fitting the model to accommodate midpoints, the final result is divided by 2 for normalisation:

\begin{equation}
\label{eq:expected_score}
    E(X_n) = \frac{1}{2} \cdot \sum_{k=0}^{K} s_k \cdot P(X_n=k)
\end{equation}

For holistic scores, instead, the final result is

\begin{equation}
\label{eq:expected_score_hol}
    E(X_n) = \sum_{k=0}^{K} s_k \cdot P(X_n=k)
\end{equation}

We refer to these final scores as \emph{fair average scores}.

Table \ref{tab:src_analysis} reports the average Spearman’s rank correlation coefficient (SRC) between the \emph{average rater} (AR)  and the reference fair average scores. In other words, we calculated the SRC between each of the 12 raters and the reference fair average scores, and reported the mean and standard deviation. However, in operational language assessment, essays are typically assigned to raters at random. To simulate this scenario, we randomly selected one rater for each essay. This procedure was repeated five times using different random seeds, and the results were averaged. We refer to this condition as the \emph{operational rater} (OR).

\begin{table}[ht!]

\centering
\begin{tabular}{lcc}
\hline
  & \textbf{AR} & \textbf{OR} \\
\hline
Intelligibility      & $0.764_{\pm .05}$   & $0.749_{\pm .02}$   \\
Complexity            & $0.770_{\pm .04}$   & $0.737_{\pm .02}$   \\
Accuracy             & $0.765_{\pm .05}$   & $0.735_{\pm .02}$   \\
Fluency              & $0.761_{\pm .03}$   & $0.733_{\pm .02}$   \\
\hline
Comprehensibility     & $0.769_{\pm .05}$   & $0.741_{\pm .02}$   \\
Logicality           & $0.748_{\pm .04}$   & $0.723_{\pm .02}$   \\
Sophistication        & $0.761_{\pm .03}$   & $0.741_{\pm .04}$   \\
Purposefulness        & $0.740_{\pm .05}$   & $0.730_{\pm .02}$   \\
\hline
Willingness          & $0.718_{\pm .03}$   & $0.696_{\pm .05}$   \\
Involvement           & $0.692_{\pm .07}$   & $0.644_{\pm .05}$   \\
\hline
\hline
Holistic              & $0.788_{\pm .04}$   & $0.775_{\pm .02}$   \\
\hline
\end{tabular}
\caption{SRC correlations of average rater (AR) and operational rater (OR) with fair average scores.}
\label{tab:src_analysis}
\end{table}

Notably, holistic scores exhibit the highest levels of agreement, as shown in Table \ref{tab:reliability_k}, a pattern that is further supported by the SRC results in Table \ref{tab:src_analysis}. This observation aligns with previous research indicating that holistic scoring is generally easier and more intuitive for human raters than analytic scoring, hence tending to yield higher inter-rater reliability than fine-grained analytic judgments~\cite{weigle2002assessing, zhang2015rater}. Furthermore, the \emph{Attitude} aspects (i.e., \emph{Willingness to communicate} and \emph{Involvement}) are inherently more subjective and consequently more difficult for raters to evaluate, as reflected in their lower SRC and Krippendorff’s $\alpha$ values compared with the other aspects.

\subsection{LLM prompting and score extraction}

We use GPT-4.1~\cite{openai2023gpt4}, Qwen 2.5 72B (4-bit quantised)~\cite{qwen2.5full}, and Llama 3.1 70B (4-bit quantised)~\cite{llama3} in a zero-shot setting to evaluate each language aspect in the essays, allowing us to compare both proprietary high-parameter and open-source, medium-size LLMs. The prompt template provided to the LLMs can be found in Appendix \ref{sec:appendix_prompt}. This contains an analytic rating prompt for each language aspect, originally used by human raters. Because the original analytic prompts~\cite[pp.~34–36]{ishikawa2024icnale} were designed for both essays and speeches, we adapted them to include only content relevant to essays. The revised prompts are provided in Appendix \ref{sec:appendix_rating_prompts}.

To extract proficiency scores from the LLM, we use a weighted average approach~\cite{ma25b_interspeech, banno25_slate}. For each analytic aspect, the LLM is prompted independently, producing a probability distribution over discrete proficiency levels via a softmax layer. Let $\mathbf{p} = [p_1, p_2, \dots, p_K]$ denote this probability distribution over $K$ ordinal levels, and let $\mathbf{v} = [v_1, v_2, \dots, v_K]$ represent the numeric values assigned to each level (here, 0 through 9).\footnote{We used a 0–9 scale because including 10 would require two tokens (``1'' and ``0''), making it difficult to extract a probability for this score.} The weighted average score for the LLM prediction is then computed as:

\begin{equation}
\text{WAvg} = \sum_{k=1}^{K} p_k \cdot v_k
\end{equation}

Intuitively, this procedure produces a continuous estimate of the score by weighting each possible level by the LLM’s predicted probability. 

\subsection{Limitations of assessing analytic scoring using rank-based metrics}
\label{sec:lim_rank}

In real-world multi-aspect language assessment, a common characteristic is the high intercorrelation among scores assessing different aspects, as well as the strong correlation between analytic and holistic scores~\cite{lee2008analytic, ono2019holistic}. This pattern is also observed in our study. Table \ref{tab:holistic_corrs} presents the SRC between predictions from various systems and the reference fair average scores for three aspects, each representing a macro-aspect: \emph{Complexity} for Language, \emph{Logicality} for Content, and \emph{Willingness} for Attitude. We compare these with the correlations between holistic fair average scores and analytic fair average scores (first row). Additionally, we report the SRC correlations on analytic fair average scores when we prompt our systems for holistic scoring.
As can be seen, holistic fair average scores correlate highly with analytic scores. As a result, when we use LLMs prompted to predict holistic scores, we observe strong correlations with analytic scores, in some cases even stronger than when we prompt them for the respective analytic aspect. For example, when prompted for holistic scores,
GPT-4.1 shows a correlation with \emph{Complexity} of 0.927 (vs 0.848) or Qwen 2.5 shows a correlation with \emph{Willingness} of 0.754 (vs 0.457). Similarly, the operational rater (OR) achieves higher correlations with analytic scores when providing holistic ratings: $0.763_{\pm .02}$ for \emph{Complexity}, $0.762_{\pm .01}$ for \emph{Logicality}, and $0.750_{\pm .02}$ for \emph{Willingness}, compared to $0.737_{\pm .02}$, $0.723_{\pm .02}$, and $0.696_{\pm .05}$, respectively, when performing analytic assessment. 
In addition to
the presence of real intercorrelations among analytic scores, this pattern may be explained by a halo effect, in which the holistic judgments influence ratings across multiple analytic dimensions, effectively acting as a proxy for the learner’s analytic ability.

Consequently, evaluating analytic assessment quality using rank-based metrics such as SRC can be misleading:\footnote{Due to the continuous nature of our scores, we cannot use QWK. However, because this also measures rank agreement, we expect it would yield similar patterns.} these metrics primarily capture global agreement in learner ranking, and may obscure the model’s ability to detect relative strengths and weaknesses within a single learner’s profile. In other words, a system can achieve high correlations with analytic scores simply by reflecting holistic proficiency rather than accurately differentiating performance across individual aspects. It is nevertheless noteworthy that LLMs outperform OR on \emph{Complexity}, \emph{Logicality}, and \emph{Holistic} (with the exception of Qwen 2.5), whereas on the more subjective dimension of \emph{Willingness}, OR still achieves the best -- albeit comparably lower -- performance, as expected (see end of Section \ref{sec:data}). The scatterplots for the three selected analytic aspects and holistic proficiency are shown in Figure \ref{fig:images_part1} in Appendix \ref{appendix:additional_stats}.

\begin{table*}[ht!] 
\centering 
\small
\setlength{\tabcolsep}{3.5pt} 

\begin{tabular}{l l| >{\raggedright\arraybackslash}p{5.1em} >{\raggedright\arraybackslash}p{5.1em} >{\raggedright\arraybackslash}p{5.1em} | >{\raggedright\arraybackslash}p{5.1em}}
\toprule
 &  & \textbf{Complexity} & \textbf{Logicality} & \textbf{Willingness} & \textbf{Holistic} \\ 
\midrule 

 & \textbf{Fair Avg. Holistic} 
 & 0.993\phantom{$_{\pm .00}$} 
 & 0.982\phantom{$_{\pm .00}$} 
 & 0.976\phantom{$_{\pm .00}$} 
 & 1.000\phantom{$_{\pm .00}$} \\ 

\midrule \midrule 

\multirow{4}{*}{\textbf{Prompted for analytic}} 
& \textbf{OR} 
& $0.737_{\pm .02}$ 
& $0.723_{\pm .02}$ 
& $0.696_{\pm .05}$ 
& - \\ 

& \textbf{GPT4.1} 
& 0.848\phantom{$_{\pm .00}$} 
& 0.810\phantom{$_{\pm .00}$} 
& 0.340\phantom{$_{\pm .00}$} 
& - \\ 

& \textbf{Qwen2.5} 
& 0.753\phantom{$_{\pm .00}$} 
& 0.769\phantom{$_{\pm .00}$} 
& 0.457\phantom{$_{\pm .00}$} 
& - \\ 

& \textbf{Llama3.1} 
& 0.814\phantom{$_{\pm .00}$} 
& 0.824\phantom{$_{\pm .00}$} 
& 0.554\phantom{$_{\pm .00}$} 
& - \\ 

\hline

\multirow{4}{*}{\textbf{Prompted for holistic}} 
& \textbf{OR} 
& $0.763_{\pm .02}$ 
& $0.762_{\pm .01}$ 
& $0.750_{\pm .02}$ 
& $0.775_{\pm .02}$  \\ 

& \textbf{GPT4.1} 
& 0.927\phantom{$_{\pm .00}$} 
& 0.921\phantom{$_{\pm .00}$}
& 0.910\phantom{$_{\pm .00}$} 
& 0.936 \\ 

& \textbf{Qwen2.5} 
& 0.738\phantom{$_{\pm .00}$}
& 0.764\phantom{$_{\pm .00}$}
& 0.754\phantom{$_{\pm .00}$} 
& 0.771  \\ 

& \textbf{Llama3.1} 
& 0.758\phantom{$_{\pm .00}$}
& 0.771\phantom{$_{\pm .00}$}
& 0.724\phantom{$_{\pm .00}$}
& 0.781  \\ 

\bottomrule 
\end{tabular} 

\caption{SRC of selected aspects (\emph{Complexity}, \emph{Logicality}, \emph{Willingness}) with fair average scores across different systems.} 
\label{tab:holistic_corrs} 
\end{table*}

The next section addresses this issue by proposing an alternative approach to analytic assessment.

\subsection{Implementing self-referential analytic assessment}

We first standardise the reference analytic fair average scores. For each analytic aspect $i$, we compute

\begin{equation}
z_i = \frac{s_i - m_i}{\sigma_i}
\end{equation}
\noindent
where $s_i$ denotes the fair average score, and $m_i$ and $\sigma_i$ are the mean and standard deviation of that score for $i$ across essays.

For each essay, we then compute the average of the standardised analytic scores across all $A$ aspects:

\begin{equation}
\mu = \frac{1}{A} \sum_{j=1}^{A} z_j
\end{equation}

We define the difference between $\mu$ and the score for aspect $i$ as

\begin{equation}
\Delta_i = \mu - z_i
\end{equation}

This difference is used to construct a dual binary classification task corresponding to negative and positive feedback. Specifically, for each aspect $i$, we define the negative feedback target as $\Delta_i \ge \sigma_{\Delta_i}$ and the positive feedback target as $\Delta_i \le -\sigma_{\Delta_i}$, where $\sigma_{\Delta_i}$ is the standard deviation of $\Delta_i$ across essays.\footnote{A one-standard-deviation threshold has also been used in the aforementioned work by \citet{berninger2010listening}.}

For model predictions, let $\hat{s}_i$ denote the predicted score for aspect $i$. Predictions are standardised analogously:

\begin{equation}
\hat{z}_i = \frac{\hat{s}_i - \hat{m}_i}{\hat{\sigma}_i}
\end{equation}
\noindent
where $\hat{m}_i$ and $\hat{\sigma}_i$ are computed over the predicted scores for aspect $i$. The average of the standardised predictions across aspects is then

\begin{equation}
\hat{\mu} = \frac{1}{A} \sum_{j=1}^{A} \hat{z}_j
\end{equation}

Finally, we compute the difference between $\hat{\mu}$ and the predicted score for aspect $i$

\begin{equation}
\hat{\Delta}_i = \hat{\mu} - \hat{z}_i
\end{equation}
\noindent
which serves as a continuous scoring signal for the feedback classification task. We report the best $F_{0.5}$ score for each aspect prioritising precision over recall to minimise false positives and avoid misleading learners about their proficiency strengths or weaknesses.

In order to be able to have a fair comparison of the results across the two binary tasks, we normalise the Precision and $F_{0.5}$ score using the method illustrated in \citet{raina2023tackling} setting the prevalence to 10\% since we are working with different language aspects, each having a different original prevalence rate (see Tables \ref{tab:feedback_random}, \ref{tab:feedback_gpt}, \ref{tab:feedback_qwen}, and \ref{tab:feedback_llama} in Appendix \ref{appendix:additional_stats}). To do this, we rescaled the proportion of negative cases so that the effective prevalence matched the target. Let $N_{+}$ and $N_{-}$ denote the number of positive and negative samples, and let $pr_{\text{target}}$ denote the desired prevalence. We first solved
\begin{equation}
\label{eq_dice}
\frac{N_{+}}{N_{+} + \gamma\,N_{-}} = pr_{\text{target}}
\end{equation}
\noindent
for the scaling factor $\gamma$ applied to the number of negatives. Assuming the classifier’s false positive rate per negative instance remains constant, false positives scale proportionally by the same factor $\gamma$. The normalised precision $P'$ is then obtained by recomputing precision using this adjusted false--positive count. Normalised recall is unaffected, and the normalised $F_{0.5}$ score is subsequently computed from $P'$ and recall $R$.

Given this prevalence value, if we calculate the $F_{0.5}$ score for a random classifier, where $\beta$ is 0.5, $P'$ is 0.10, and $R$ is 1:

\begin{equation}
    F_\beta = (1 + \beta^2)\,\frac{\text{$P'$} \cdot \text{$R$}}{\beta^2 \cdot \text{$P'$} + \text{$R$}}
\end{equation}

\noindent
we obtain a score of $\sim12.19$. Any results higher than this value contain information. Table \ref{tab:feed_percentage} in Appendix \ref{appendix:additional_stats} illustrates the percentage of essays receiving feedback across macro-aspects.

Before discussing the results, it is important to clarify the nature of the task. Unlike the inter-learner correlation analyses reported in Table \ref{tab:holistic_corrs}, our proposed self-referential framework focuses on identifying aspects that are unusually weak or strong relative to an individual learner’s overall profile. As a result, systems may show high correlations while still differing substantially in how they distribute strengths and weaknesses across analytic aspects within the same essay.

\section{Experimental results}

Tables \ref{tab:negative-feedback} and \ref{tab:positive-feedback} show the results for GPT-4.1, Qwen 2.5 72B, Llama 3.1 70B, and OR on negative and positive feedback, respectively.

\begin{table}[ht!]
\small
\centering
\setlength{\tabcolsep}{3pt}
\begin{tabular}{@{}lcccc@{}}
\hline
 & \textbf{GPT4.1} & \textbf{Qwen2.5} & \textbf{Llama3.1} & \textbf{OR} \\
\hline
Int & \textbf{44.68} & 31.96 & 40.92 & $40.77_{\pm 15.37}$ \\
Cpl & \textbf{49.20} & 37.20 & 34.00 & $30.76_{\pm 8.01}$ \\
Acc & 24.68 & 16.75 & 24.66 & \bm{$26.10_{\pm 3.80}$} \\
Flu & \textbf{22.52} & 17.70 & 20.51 & $22.37_{\pm 5.45}$ \\
\hline
Cpr & \textbf{37.71} & 15.71 & 25.55 & $34.10_{\pm 7.78}$ \\
Lgc & \textbf{45.45} & 36.52 & 28.18 & $23.67_{\pm 7.67}$ \\
Sph & 20.52 & \textbf{27.71} & 20.20 & $23.83_{\pm 6.11}$ \\
Prp & 38.20 & 46.13 & \textbf{51.43} & $33.25_{\pm 9.22}$ \\
\hline
Wil & 16.37 & 12.72 & 20.11 & \bm{$32.79_{\pm 7.61}$} \\
Inv & 37.75 & 40.51 & 42.93 & \bm{$45.87_{\pm 4.38}$} \\
\hline\hline
Avg. & \textbf{33.71} & 28.29 & 30.85 & 31.35 \\
\hline
\end{tabular}
\caption{\textbf{Negative} feedback results in terms of best $F_{0.5}$ across systems.}
\label{tab:negative-feedback}
\end{table}

\begin{table}[ht!]
\small
\centering
\setlength{\tabcolsep}{3pt}
\begin{tabular}{@{}lcccc@{}}
\hline
 & \textbf{GPT4.1} & \textbf{Qwen2.5} & \textbf{Llama3.1} & \textbf{OR} \\
\hline
Int & 21.02 & 13.80 & 22.84 & \bm{$35.59_{\pm 5.34}$} \\
Cpl & \textbf{74.07} & 65.22 & 70.00 & $41.58_{\pm 6.13}$ \\
Acc & 29.96 & 33.61 & 28.19 & \bm{$34.75_{\pm 8.60}$} \\
Flu & 19.44 & 18.69 & 22.38 & \bm{$30.23_{\pm 4.04}$} \\
\hline
Cpr & 23.65 & 14.11 & 23.74 & \bm{$29.75_{\pm 3.13}$} \\
Lgc & 21.18 & 37.63 & \textbf{40.52} & $33.06_{\pm 3.64}$ \\
Sph & 37.74 & \textbf{39.47} & 35.71 & $29.51_{\pm 6.83}$ \\
Prp & 25.68 & 27.23 & 37.64 & \bm{$38.23_{\pm 5.27}$} \\
\hline
Wil & 29.25 & 29.71 & 26.89 & \bm{$32.81_{\pm 11.11}$} \\
Inv & 37.68 & 38.98 & \textbf{47.70} & $40.24_{\pm 7.25}$ \\
\hline\hline
Avg. & 31.96 & 31.84 & \textbf{35.56} & 34.57 \\
\hline
\end{tabular}
\caption{\textbf{Positive} feedback results in terms of best $F_{0.5}$ across systems.}
\label{tab:positive-feedback}
\end{table}

As can be observed, when examining individual analytic aspects under the negative feedback condition (Table \ref{tab:negative-feedback}), LLMs, particularly GPT-4.1, tend to achieve higher $F_{0.5}$ scores than the OR on several \emph{Language} and \emph{Content} dimensions. GPT-4.1 notably outperforms both the other models and OR on \emph{Intelligibility}, \emph{Complexity}, \emph{Fluency}, \emph{Comprehensibility}, and \emph{Logicality}, indicating a stronger alignment with the reference signal in identifying aspects that deviate negatively compared to the overall average score. This suggests that LLMs are especially effective at detecting relative weaknesses in linguistically and structurally grounded dimensions. By contrast, OR attains higher scores on \emph{Accuracy} and on the \emph{Attitude} dimensions (i.e., \emph{Willingness} and \emph{Involvement}), for which all LLMs show comparatively lower performance. Given that \emph{Attitude} aspects are inherently more subjective and have been shown to exhibit lower inter-rater agreement (see Tables \ref{tab:reliability_k} and \ref{tab:src_analysis}), these differences likely reflect increased noise and variability in the reference labels rather than a systematic limitation of the self-referential assessment formulation or the models themselves. Within the group of LLMs, GPT-4.1 consistently leads across most \emph{Language} and \emph{Content} aspects, while Qwen 2.5 performs best on \emph{Sophistication}, and Llama 3.1 achieves its highest scores on \emph{Purposefulness}. However, overall, GPT-4.1 attains the highest average $F_{0.5}$ score for negative feedback.

For positive feedback (Table \ref{tab:positive-feedback}), a different pattern is observed. While LLMs continue to perform strongly on \emph{Complexity}, \emph{Logicality}, and \emph{Sophistication}, the OR outperforms all models on most \emph{Language} aspects, including \emph{Intelligibility}, \emph{Accuracy}, and \emph{Fluency}, as well as on \emph{Comprehensibility} and \emph{Purposefulness}. Among the LLMs, Llama 3.1 achieves the highest overall average $F_{0.5}$ score under positive feedback, driven primarily by its performance on \emph{Logicality}, \emph{Purposefulness}, and \emph{Involvement}. In contrast, GPT-4.1, while dominant in the negative feedback setting, shows comparatively weaker alignment with the positive feedback targets.
Overall, we can observe that LLMs, especially GPT-4.1, exhibit stronger alignment with reference signals for identifying relative underperformance, whereas human raters show higher agreement with the reference in identifying relative strengths. For completeness, in addition to $F_{0.5}$, we also report the results in terms of Precision and Recall in Appendix \ref{appendix:additional_stats} in Tables \ref{tab:feedback_random}, \ref{tab:feedback_gpt}, \ref{tab:feedback_qwen}, and \ref{tab:feedback_llama} for OR, GPT-4.1, Qwen 2.5, and Llama 3.1, respectively.

\begin{figure}[ht!]
\centering

\begin{subfigure}{0.482\textwidth}
\centering
\includegraphics[width=\linewidth]{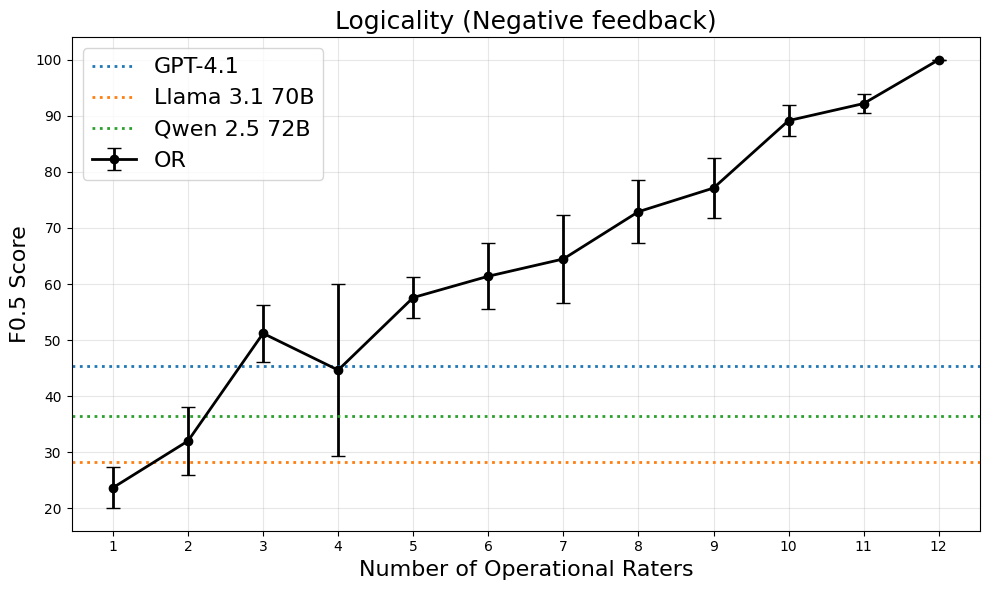}
\end{subfigure}
\hfill
\begin{subfigure}{0.482\textwidth}
\centering
\includegraphics[width=\linewidth]{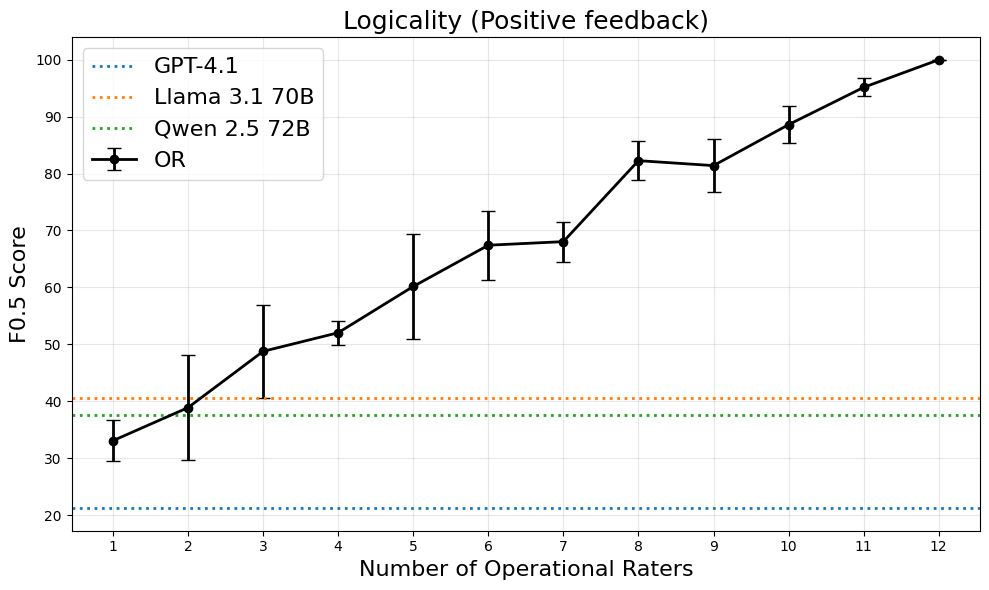}

\end{subfigure}

\caption{Errorbars on feedback of ensembles with increasing number of operational raters to GPT-4.1, Llama 3.1 70B, and Qwen 2.5 72B for \emph{Logicality}.}
\label{errorbar_lgc}
\end{figure}

In addition to comparing LLMs with a single randomly selected operational rater per essay, we also report performance for ensembles of multiple raters, ranging in size from 2 to 12. Figure \ref{errorbar_lgc} reports the errorbars for feedback on \emph{Logicality}. As can be seen, for this aspect, it takes an ensemble of three raters to outperform the best-performing models, i.e., GPT-4.1 on negative feedback and Llama 3.1 on positive feedback.

For completeness, we report the results for all the aspects and one, two, and three operational raters in Tables \ref{tab:ors_results_negative} and \ref{tab:ors_results_positive} in Appendix \ref{appendix:additional_stats}. When averaging performance across aspects, an ensemble of two operational raters is sufficient to outperform the LLMs. However, considering individual aspects, as in Figure \ref{errorbar_lgc}, larger ensembles are required.

\section{Conclusions}

This paper addresses the challenges of validating diagnostic analytic assessment in AES, particularly where rank-based metrics may overlook the influence of intercorrelations among analytic dimensions of proficiency as well as the halo effect. Through the introduction of a self-referential framework, we demonstrate how intra-learner evaluation serves as a crucial complement to standard inter-learner assessment.

Our results highlight divergent strengths: LLMs tend to outperform operational human raters in pinpointing relative weaknesses (negative feedback), while humans remain more adept at recognising relative strengths (positive feedback). While the primary focus of this paper is the proposed self-referential assessment framework, we acknowledge that our experiments are limited to zero-shot approaches. Future work will investigate alternative prompting strategies, including few-shot prompting and chain-of-thought reasoning.

Finally, while inter-learner rankings provide essential comparative data, we propose combining them with intra-learner diagnostics. This hybrid approach may offer a pathway to delivering more actionable and informative assessment and feedback to learners, teachers, and testers.

\section*{Limitations}

First, we acknowledge that relying on a single dataset warrants caution; further experiments are therefore necessary to fully validate the effectiveness of our self-referential assessment framework. While the number of unique essays is relatively small ($N=140$), the original dataset is exceptionally dense, containing over 120,000 ratings from 80 raters. This unique characteristic of the ICNALE GRA allowed us to implement Rasch modelling in a fully crossed rater–essay design, a rigorous approach not feasible with other L2 learner datasets.

Secondly, as the data focuses on Asian learners within an English as a Lingua Franca (ELF) framework, further research is required to determine if these findings generalise to learners from other L1 backgrounds or for other L2s.

Finally, regarding the self-referential framework, while the use of a one-standard-deviation threshold is a heuristic choice, it ensures a consistent data-driven approach and aligns with established methodology in psychometric literature~\cite{berninger2010listening}. Moreover, it guarantees that a substantial proportion of learners receive both positive and negative feedback, as shown in Table \ref{tab:feed_percentage} (Appendix \ref{appendix:additional_stats}).

\section*{Acknowledgments}

This paper reports on research supported by Cambridge University Press \& Assessment, a department of The Chancellor, Masters, and Scholars of the University of Cambridge. The authors would like to thank the ALTA Spoken Language Processing Technology Project Team for general discussions and contributions to the evaluation infrastructure.


\bibliography{custom}

\begin{thebibliography}{47}
\providecommand{\natexlab}[1]{#1}

\bibitem[{Bannò et~al.(2025)Bannò, Ma, Qian, Tang, Knill, and Gales}]{banno25_slate}
Stefano Bannò, Rao Ma, Mengjie Qian, Siyuan Tang, Kate Knill, and Mark Gales. 2025.
\newblock \href {https://doi.org/10.21437/SLaTE.2025-38} {{Natural Language-based Assessment of L2 Oral Proficiency using LLMs}}.
\newblock In \emph{{10th Workshop on Speech and Language Technology in Education (SLaTE)}}, pages 189--193.

\bibitem[{Bannò et~al.(2024)Bannò, Vydana, Knill, and Gales}]{banno-etal-2024-gptfull}
Stefano Bannò, Hari~Krishna Vydana, Kate Knill, and Mark Gales. 2024.
\newblock \href {https://aclanthology.org/2024.bea-1.14} {Can {GPT}-4 do {L}2 analytic assessment?}
\newblock In \emph{Proc. of the 19th Workshop on Innovative Use of NLP for Building Educational Applications (BEA 2024)}, pages 149--164, Mexico City, Mexico. Association for Computational Linguistics.

\bibitem[{Berninger and Abbott(2010)}]{berninger2010listening}
Virginia~W Berninger and Robert~D Abbott. 2010.
\newblock Listening comprehension, oral expression, reading comprehension, and written expression: Related yet unique language systems in grades 1, 3, 5, and 7.
\newblock \emph{Journal of educational psychology}, 102(3):635.

\bibitem[{Cattell(1944)}]{cattell1944psychological}
Raymond~B Cattell. 1944.
\newblock Psychological measurement: normative, ipsative, interactive.
\newblock \emph{Psychological review}, 51(5):292.

\bibitem[{Clemans(1956)}]{clemans1956analytical}
William~Vance Clemans. 1956.
\newblock \emph{{An analytical and empirical examination of some properties of ipsative measures (Psychometric Monograph No. 14)}}.
\newblock Psychometric Society, Richmond, VA.

\bibitem[{{Council of Europe}(2020)}]{cefr2020}
{Council of Europe}. 2020.
\newblock \emph{Common European Framework of Reference for Languages: Learning, Teaching, Assessment - Companion volume}.
\newblock Council of Europe, Strasbourg.

\bibitem[{Cronbach(1951)}]{cronbach1951coefficient}
Lee~J Cronbach. 1951.
\newblock Coefficient alpha and the internal structure of tests.
\newblock \emph{Psychometrika}, 16(3):297--334.

\bibitem[{Do et~al.(2023)Do, Kim, and Lee}]{do-etal-2023-prompt}
Heejin Do, Yunsu Kim, and Gary~Geunbae Lee. 2023.
\newblock \href {https://doi.org/10.18653/v1/2023.findings-acl.98} {{Prompt- and Trait Relation-aware Cross-prompt Essay Trait Scoring}}.
\newblock In \emph{Findings of the Association for Computational Linguistics: ACL 2023}, pages 1538--1551, Toronto, Canada. Association for Computational Linguistics.

\bibitem[{Engelhard(1994)}]{engelhard1994examining}
George Engelhard. 1994.
\newblock \href {http://www.jstor.org/stable/1435170} {E{xamining Rater Errors in the Assessment of Written Composition with a Many-Faceted Rasch Model}}.
\newblock \emph{Journal of Educational Measurement}, 31(2):93--112.

\bibitem[{Hughes(2011)}]{hughes2011towards}
Gwyneth Hughes. 2011.
\newblock Towards a personal best: A case for introducing ipsative assessment in higher education.
\newblock \emph{Studies in Higher Education}, 36(3):353--367.

\bibitem[{Hussein et~al.(2020)Hussein, Hassan, and Nassef}]{hussein2020trait-based}
Mohamed~A. Hussein, Hesham~A. Hassan, and Mohammad Nassef. 2020.
\newblock \href {https://doi.org/10.14569/IJACSA.2020.0110538} {A trait-based deep learning automated essay scoring system with adaptive feedback}.
\newblock \emph{International Journal of Advanced Computer Science and Applications}, 11(5).

\bibitem[{Ishikawa(2011)}]{ishikawa2011}
Shin'Ichiro Ishikawa. 2011.
\newblock A new horizon in learner corpus studies: The aim of the {ICNALE} project.
\newblock In \emph{Corpora and language technologies in teaching, learning and research}, pages 3--11. University of Strathclyde Press.

\bibitem[{Ishikawa(2020)}]{ishikawa2020}
Shin'Ichiro Ishikawa. 2020.
\newblock {Aim of the {ICNALE} {GRA} Project: Global Collaboration to Collect Ratings of Asian Learners' L2 English Essays and Speeches from an {ELF} Perspective}.
\newblock \emph{Learner Corpus Studies in Asia and the World}, 5:121--144.

\bibitem[{Ishikawa(2024)}]{ishikawa2024icnale}
Shin'Ichiro Ishikawa. 2024.
\newblock {The ICNALE Global Rating Archives: A New Assessment Dataset for Learner Corpus Studies}.
\newblock \emph{Learner Corpus Studies in Asia and the World}, 6:13--38.

\bibitem[{Ishikawa(2023)}]{ishikawa2023effects}
Shin’Ichiro Ishikawa. 2023.
\newblock {Effects of Raters’ L1, Assessment Experience, and Teaching Experience on their Assessment of L2 English Speech: A Study Based on the ICNALE Global Rating Archives}.
\newblock \emph{LEARN Journal: Language Education and Acquisition Research Network}, 16(2):411--428.

\bibitem[{Klebanov and Madnani(2022)}]{klebanov2022automated}
Beata~Beigman Klebanov and Nitin Madnani. 2022.
\newblock \href {https://doi.org/10.1007/978-3-031-02182-4} {\emph{Automated Essay Scoring}}.
\newblock Springer Nature.

\bibitem[{Krippendorff(2004)}]{krippendorff2004reliability}
Klaus Krippendorff. 2004.
\newblock \href {https://doi.org/10.1111/j.1468-2958.2004.tb00738.x} {{Reliability in Content Analysis}}.
\newblock \emph{Human Communication Research}, 30(3):411--433.

\bibitem[{Krippendorff(2011)}]{krippendorff2011computing}
Klaus Krippendorff. 2011.
\newblock \href {https://www.asc.upenn.edu/sites/default/files/2021-03/Computing%20Krippendorff%27s%20Alpha-Reliability.pdf} {{Computing Krippendorff's Alpha-Reliability}}.

\bibitem[{Kumar et~al.(2022)Kumar, Mathias, Saha, and Bhattacharyya}]{kumar-etal-2022-many}
Rahul Kumar, Sandeep Mathias, Sriparna Saha, and Pushpak Bhattacharyya. 2022.
\newblock \href {https://doi.org/10.18653/v1/2022.naacl-main.106} {{Many Hands Make Light Work: Using Essay Traits to Automatically Score Essays}}.
\newblock In \emph{Proceedings of the 2022 Conference of the North American Chapter of the Association for Computational Linguistics: Human Language Technologies}, pages 1485--1495, Seattle, United States. Association for Computational Linguistics.

\bibitem[{Lee et~al.(2008)Lee, Gentile, and Kantor}]{lee2008analytic}
Yong-Won Lee, Claudia Gentile, and Robert Kantor. 2008.
\newblock {Analytic scoring of TOEFL{\textregistered} CBT essays: Scores from humans and e-rater{\textregistered}}.
\newblock \emph{ETS Research Report Series}, 2008(1):i--71.

\bibitem[{Li and Ng(2024)}]{li2024essay}
Shengjie Li and Vincent Ng. 2024.
\newblock \href {https://doi.org/10.24963/ijcai.2024/897} {Automated essay scoring: Recent successes and future directions}.
\newblock In \emph{Proc. of the Thirty-Third International Joint Conference on Artificial Intelligence, {IJCAI-24}}, pages 8114--8122.
\newblock {Survey Track}.

\bibitem[{Linacre(1989)}]{linacre1989many}
John~Michael Linacre. 1989.
\newblock \emph{{Many-faceted Rasch measurement}}.
\newblock Ph.D. thesis, The University of Chicago.

\bibitem[{Linacre(2002)}]{linacre2002infit}
John~Michael Linacre. 2002.
\newblock What do infit and outfit, mean-square and standardized mean.
\newblock \emph{Rasch measurement transactions}, 16(2):878.

\bibitem[{{Llama Team}(2024)}]{llama3}
{Llama Team}. 2024.
\newblock \href {https://ai.meta.com/research/publications/the-llama-3-herd-of-models/} {The {Llama} 3 herd of models}.

\bibitem[{Lombard et~al.(2006)Lombard, Snyder-Duch, and Bracken}]{lombard2006content}
Matthew Lombard, Jennifer Snyder-Duch, and Cheryl~Campanella Bracken. 2006.
\newblock \href {https://doi.org/10.1111/j.1468-2958.2002.tb00826.x} {{Content Analysis in Mass Communication: Assessment and Reporting of Intercoder Reliability}}.
\newblock \emph{Human Communication Research}, 28(4):587--604.

\bibitem[{Ma et~al.(2025)Ma, Qian, Tang, Bannò, Knill, and Gales}]{ma25b_interspeech}
Rao Ma, Mengjie Qian, Siyuan Tang, Stefano Bannò, Kate~M. Knill, and Mark~J.F. Gales. 2025.
\newblock \href {https://doi.org/10.21437/Interspeech.2025-1793} {{Assessment of L2 Oral Proficiency using Speech Large Language Models}}.
\newblock In \emph{{Interspeech 2025}}, pages 5078--5082.

\bibitem[{Mathias and Bhattacharyya(2020)}]{mathias-bhattacharyya-2020-neural}
Sandeep Mathias and Pushpak Bhattacharyya. 2020.
\newblock \href {https://doi.org/10.18653/v1/2020.bea-1.8} {{Can Neural Networks Automatically Score Essay Traits?}}
\newblock In \emph{Proc. of the Fifteenth Workshop on Innovative Use of NLP for Building Educational Applications}, pages 85--91. Association for Computational Linguistics.

\bibitem[{Mizumoto and Eguchi(2023)}]{mizumoto2023gpt}
Atsushi Mizumoto and Masaki Eguchi. 2023.
\newblock \href {https://doi.org/10.1016/j.rmal.2023.100050} {{Exploring the potential of using an AI language model for automated essay scoring}}.
\newblock \emph{Research Methods in Applied Linguistics}, 2(2):100050.

\bibitem[{Naismith et~al.(2023)Naismith, Mulcaire, and Burstein}]{naismith-etal-2023-automatedfull}
Ben Naismith, Phoebe Mulcaire, and Jill Burstein. 2023.
\newblock \href {https://doi.org/10.18653/v1/2023.bea-1.32} {Automated evaluation of written discourse coherence using {GPT}-4}.
\newblock In \emph{Proc. of the 18th Workshop on Innovative Use of NLP for Building Educational Applications (BEA 2023)}, pages 394--403, Toronto, Canada. Association for Computational Linguistics.

\bibitem[{Ono et~al.(2019)Ono, Yamanishi, and Hijikata}]{ono2019holistic}
Masumi Ono, Hiroyuki Yamanishi, and Yuko Hijikata. 2019.
\newblock {Holistic and analytic assessments of the TOEFL iBT{\textregistered} Integrated Writing Task}.
\newblock \emph{JLTA Journal}, 22:65--88.

\bibitem[{OpenAI(2023)}]{openai2023gpt4}
OpenAI. 2023.
\newblock \href {https://doi.org/10.48550/arXiv.2303.08774} {{GPT-4 Technical Report}}.
\newblock \emph{Preprint}, arXiv:2303.08774.

\bibitem[{Page(1966)}]{page1966imminence}
Ellis~B. Page. 1966.
\newblock The imminence of... grading essays by computer.
\newblock \emph{The Phi Delta Kappan}, 47(5):238--243.

\bibitem[{Page et~al.(1997)Page, Poggio, and Keith}]{page1997computer}
Ellis~B Page, John~P Poggio, and Timothy~Z Keith. 1997.
\newblock {Computer Analysis of Student Essays: Finding Trait Differences in Student Profile}.
\newblock In \emph{Annual Meeting of the American Educational Research Association}, Chicago, IL.

\bibitem[{Raina et~al.(2023)Raina, Molchanova, Graziani, Malinin, Muller, Cuadra, and Gales}]{raina2023tackling}
Vatsal Raina, Nataliia Molchanova, Mara Graziani, Andrey Malinin, Henning Muller, Meritxell~Bach Cuadra, and Mark Gales. 2023.
\newblock {Tackling Bias in the Dice Similarity Coefficient: Introducing nDSC for White Matter Lesion Segmentation}.
\newblock In \emph{2023 IEEE 20th International Symposium on Biomedical Imaging (ISBI)}, pages 1--5. IEEE.

\bibitem[{Ridley et~al.(2021)Ridley, He, Dai, Huang, and Chen}]{ridley2021}
Robert Ridley, Liang He, Xin-yu Dai, Shujian Huang, and Jiajun Chen. 2021.
\newblock \href {https://doi.org/10.1609/aaai.v35i15.17620} {Automated cross-prompt scoring of essay traits}.
\newblock \emph{Proc. of the AAAI Conference on Artificial Intelligence}, 35(15):13745--13753.

\bibitem[{Seidlhofer(2005)}]{seidlhofer2005elf}
Barbara Seidlhofer. 2005.
\newblock \href {https://doi.org/10.1093/elt/cci064} {English as a lingua franca}.
\newblock \emph{ELT Journal}, 59(4):339--341.

\bibitem[{Se{\ss}ler et~al.(2025)Se{\ss}ler, F{\"u}rstenberg, B{\"u}hler, and Kasneci}]{sessler2025can}
Kathrin Se{\ss}ler, Maurice F{\"u}rstenberg, Babette B{\"u}hler, and Enkelejda Kasneci. 2025.
\newblock {Can AI grade your essays? A comparative analysis of large language models and teacher ratings in multidimensional essay scoring}.
\newblock In \emph{Proceedings of the 15th International Learning Analytics and Knowledge Conference}, pages 462--472.

\bibitem[{Shermis et~al.(2002)Shermis, Koch, Page, Keith, and Harrington}]{shermis2002trait}
Mark~D Shermis, Chantal~Mees Koch, Ellis~B Page, Timothy~Z Keith, and Susanmarie Harrington. 2002.
\newblock Trait ratings for automated essay grading.
\newblock \emph{Educational and Psychological Measurement}, 62(1):5--18.

\bibitem[{Thorndike(1920)}]{thorndike1920constant}
Edward~L Thorndike. 1920.
\newblock A constant error in psychological ratings.
\newblock \emph{Journal of applied psychology}, 4(1):25--29.

\bibitem[{Wang et~al.(2025)Wang, Makarova, Li, Kodner, and Rambow}]{wang-etal-2025-llms-perform}
Zhengxiang Wang, Veronika Makarova, Zhi Li, Jordan Kodner, and Owen Rambow. 2025.
\newblock \href {https://doi.org/10.18653/v1/2025.acl-long.423} {{{LLM}s can Perform Multi-Dimensional Analytic Writing Assessments: A Case Study of {L}2 Graduate-Level Academic {E}nglish Writing}}.
\newblock In \emph{Proceedings of the 63rd Annual Meeting of the Association for Computational Linguistics (Volume 1: Long Papers)}, pages 8637--8663, Vienna, Austria. Association for Computational Linguistics.

\bibitem[{Weigle(2002)}]{weigle2002assessing}
Sara~Cushing Weigle. 2002.
\newblock \href {https://doi.org/10.1017/CBO9780511732997} {\emph{Assessing Writing}}.
\newblock Cambridge Language Assessment. Cambridge University Press, Cambridge.

\bibitem[{Yamashita(2024)}]{yamashita2024application}
Taichi Yamashita. 2024.
\newblock \href {https://doi.org/10.1016/j.rmal.2024.100133} {{An application of many-facet Rasch measurement to evaluate automated essay scoring: A case of ChatGPT-4.0}}.
\newblock \emph{Research Methods in Applied Linguistics}, 3(3):100133.

\bibitem[{Yancey et~al.(2023)Yancey, Laflair, Verardi, and Burstein}]{yancey-etal-2023-ratingfull}
Kevin~P. Yancey, Geoffrey Laflair, Anthony Verardi, and Jill Burstein. 2023.
\newblock \href {https://doi.org/10.18653/v1/2023.bea-1.49} {Rating short {L}2 essays on the {CEFR} scale with {GPT}-4}.
\newblock In \emph{Proc. of the 18th Workshop on Innovative Use of NLP for Building Educational Applications (BEA 2023)}, pages 576--584, Toronto, Canada. Association for Computational Linguistics.

\bibitem[{Yang et~al.(2024)Yang, Yang, Zhang, Hui, Zheng, Yu, Li, Liu, Huang, Wei, Lin, Yang, Tu, Zhang, Yang, Yang, Zhou, Lin, Dang, Lu, Bao, Yang, Yu, Li, Xue, Zhang, Zhu, Men, Lin, Li, Xia, Ren, Ren, Fan, Su, Zhang, Wan, Liu, Cui, Zhang, and Qiu}]{qwen2.5full}
An~Yang, Baosong Yang, Beichen Zhang, Binyuan Hui, Bo~Zheng, Bowen Yu, Chengyuan Li, Dayiheng Liu, Fei Huang, Haoran Wei, Huan Lin, Jian Yang, Jianhong Tu, Jianwei Zhang, Jianxin Yang, Jiaxi Yang, Jingren Zhou, Junyang Lin, Kai Dang, and 22 others. 2024.
\newblock \href {https://doi.org/10.48550/arXiv.2412.15115} {{Qwen2.5 Technical Report}}.
\newblock \emph{arXiv preprint arXiv:2412.15115}.

\bibitem[{Yannakoudakis and Cummins(2015)}]{yannakoudakis-cummins-2015-evaluating}
Helen Yannakoudakis and Ronan Cummins. 2015.
\newblock \href {https://doi.org/10.3115/v1/W15-0625} {{Evaluating the performance of Automated Text Scoring systems}}.
\newblock In \emph{Proceedings of the Tenth Workshop on Innovative Use of {NLP} for Building Educational Applications}, pages 213--223, Denver, Colorado. Association for Computational Linguistics.

\bibitem[{Yoo et~al.(2025)Yoo, Han, Ahn, and Oh}]{yoo-etal-2025-dress}
Haneul Yoo, Jieun Han, So-Yeon Ahn, and Alice Oh. 2025.
\newblock \href {https://doi.org/10.18653/v1/2025.acl-long.659} {{{DRE}s{S}: Dataset for Rubric-based Essay Scoring on {EFL} Writing}}.
\newblock In \emph{Proceedings of the 63rd Annual Meeting of the Association for Computational Linguistics (Volume 1: Long Papers)}, pages 13439--13454, Vienna, Austria. Association for Computational Linguistics.

\bibitem[{Zhang et~al.(2015)Zhang, Xiao, and Luo}]{zhang2015rater}
Bo~Zhang, Yunnan Xiao, and Juan Luo. 2015.
\newblock Rater reliability and score discrepancy under holistic and analytic scoring of second language writing.
\newblock \emph{Language Testing in Asia}, 5(1):5.

\end{thebibliography}

\appendix

\section{Appendix A: Prompt}
\label{sec:appendix_prompt}

\begin{quote}
    \texttt{In the context of an examination of English as a Lingua Franca (ELF), a second language (L2) learner of English is asked to write an essay in response to the following prompt:}

\texttt{Do you agree or disagree with the following statement? Use reasons and specific details to support your opinion.}

\texttt{`It is important for college students to have a part-time job.'}

\texttt{You have to score this essay by only considering the aspect of [ASPECT].}

\texttt{[ANALYTIC RATING PROMPT]\footnote{See Appendix \ref{sec:appendix_rating_prompts}.}}

\texttt{Select a score from  0 (lowest) to 9 (highest). Only output the most suitable score without adding any comment or explanation.}

\texttt{Essay: [ESSAY]}
\end{quote}

\section{Appendix B: Analytic rating prompts}
\label{sec:appendix_rating_prompts}
Since ICNALE GRA contains essays and speeches, the original analytic rating prompts \cite{ishikawa2024icnale} addressed both modalities. For our experiments with LLMs, we therefore tailored the prompts to retain only the content relevant to essays.

\subsection*{Language}

\subsubsection*{Intelligibility}

\emph{To which extent can you “decode”, namely, verbally understand what is written? Factors such as spelling and sentence structure may influence it. Please note that intelligibility, which concerns the understandability of the language, should be discriminated from comprehensibility, which concerns the understandability of the content. You may sometimes find an essay that is intelligible but not comprehensible, such as a logically nonsense statement. Meanwhile, you may usually not find an essay that is comprehensible but not intelligible because if the text cannot be decoded, its content cannot be conveyed.}

\subsubsection*{Complexity}

\emph{To what extent do you think the writer uses morphologically and/or semantically complex words, phrases, expressions, constructions, and grammar? Complexity is seen at many levels of language. For example, “I speculate...” usually sounds more complex than “I think” (Vocabulary). “It is speculated that...” may sound more complex than “I speculate” (Voice, Construction). “If I were a bird” may sound more complex than “If I am a bird” (Subjunctive, Grammar).}

\subsubsection*{Accuracy}

\emph{To what extent do you think the sample is error-free in terms of vocabulary and grammar? In addition, you should examine the elements such as punctuation. Please note that you should ignore minor and only-once errors, which may be mistakes rather than errors. Please note that the standard for evaluation should be a proficient non-native ELF speaker, not an English native speaker.}

\subsubsection*{Fluency}

\emph{To what extent do you think the writer is fluent in the essays? Fluency needs to be evaluated in two ways: (a) fluency and (b) disfluency. If someone writes more, the fluency score should increase, while if s/he uses more disfluency markers, the score may decrease. Disfluency markers include unnecessary connectors (and, but, so because) and semantically empty phrases (such as “I think” most typically), etc. Please note that using these disfluency markers once or twice usually does not cause any problems in communication.}

\subsection*{Content}

\subsubsection*{Comprehensibility}

\emph{To what extent can you understand the content of the essay? Please note that comprehensibility, which concerns the understandability of the content, should be discriminated from intelligibility, which concerns the understandability of the language. If a writer presents a logically reasonable idea, the score should increase.}

\subsubsection*{Logicality}

\emph{To what extent do you think the idea presented in the essay is logical and reasonable? You need to examine whether the reasons and the conclusions are logically connected.}

\subsubsection*{Sophistication}

\emph{To what extent do you think the ideas presented in the essay are well-sophisticated, critically thought, unique, original, and innovative?}

\subsubsection*{Purposefulness}

\emph{To what extent do you think the writer consistently and consciously pays attention to the purpose of the task? The participant was requested to persuade a supervisor to allow them to show their own opinion about part-time jobs for college students in an essay. You have to examine whether the participant fully understands the purpose of the task and consistently sticks to it. Purposefulness is closely related to task completion.}

\subsection*{Attitude}

\subsubsection*{Willingness to communicate}

\emph{To what extent do you think the writer is willing to communicate? It is possible that a participant with a limited L2 proficiency shows a high level of willingness to communicate (WTC), and it is also possible that a participant with a high L2 proficiency shows quite a low level of WTC. Factors such as the quantity of writing, the number of ideas s/he presents, and the use of various amplifiers (e.g., “very,” “surely,” “definitely,” “I strongly believe,” etc.) may represent the participant’s WTC.}

\subsubsection*{Involvement}

\emph{To what extent do you think the participant tries to make the reader involved in his/her discourse rather than writing one-sidedly? The factors such as the use of the second-person pronouns (e.g., “You know,” “as you see,” “as you expect,” etc.) and mentioning the reader are usually related to the degree of involvement.}

\subsection*{Holistic}

\emph{To what extent do you think the sample is close to an ideal ELF essay? Raters have to examine each sample and decide the score (0-100) based on the overall judgment of its quality as a professional ELF output. Please note that “9 points”,  for example, should be given to someone who you think is a 100\% ideal professional ELF user, not to someone who you think is 100\% close to English native speakers. Also, please note that the middle point is 5.}

\section{Appendix C: Additional statistics and results}
\label{appendix:additional_stats}

Table \ref{tab:feed_percentage} reports the percentage of essays receiving positive and negative feedback across macro-aspects.

\begin{figure}[ht!]
\centering

\begin{subfigure}{\linewidth}
  \centering
  \includegraphics[width=0.75\linewidth]{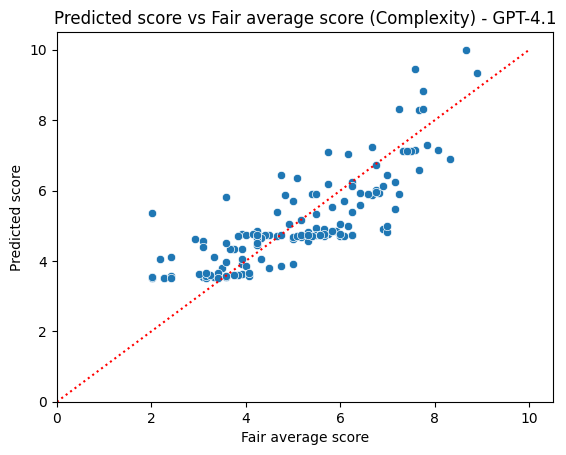}
\end{subfigure}

\vspace{0.5em}

\begin{subfigure}{\linewidth}
  \centering
  \includegraphics[width=0.75\linewidth]{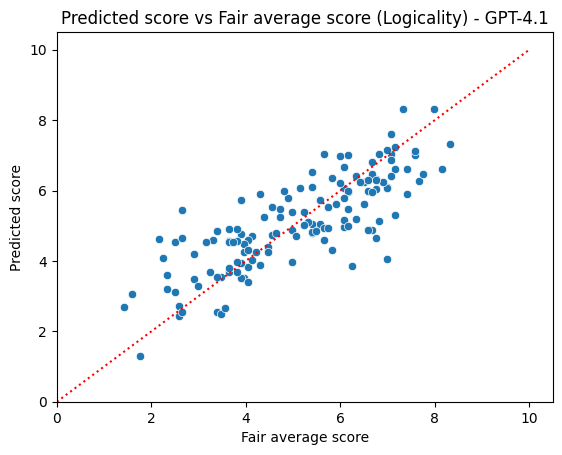}
\end{subfigure}

\vspace{0.5em}

\begin{subfigure}{\linewidth}
  \centering
  \includegraphics[width=0.75\linewidth]{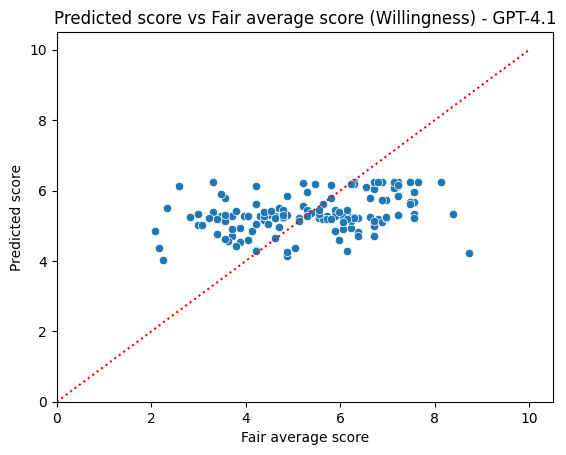}
\end{subfigure}

\vspace{0.5em}

\begin{subfigure}{\linewidth}
  \centering
  \includegraphics[width=0.75\linewidth]{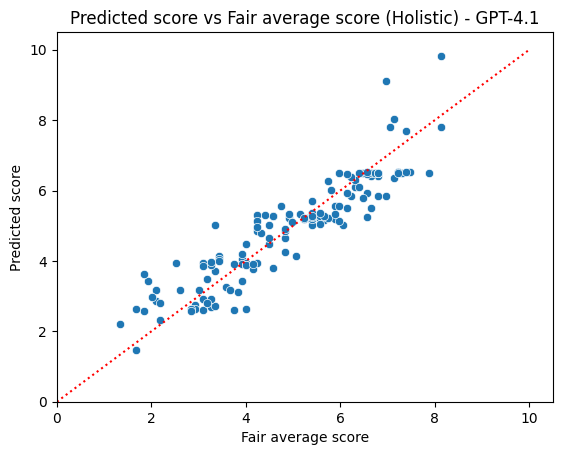}
\end{subfigure}

\caption{Scatterplots between fair average scores and GPT-4.1 scores for selected aspects.}
\label{fig:images_part1}
\end{figure}

Figure \ref{fig:images_part1} shows scatterplots comparing GPT-4.1 predictions with reference fair average scores for the three selected analytic aspects and holistic proficiency. For visualisation purposes, GPT-4.1’s predictions were linearly calibrated using the same essay test data.

Tables \ref{tab:feedback_random}, \ref{tab:feedback_gpt}, \ref{tab:feedback_qwen}, and \ref{tab:feedback_llama} show the complete feedback results in terms of Precision, Recall, and $F_{0.5}$ for OR, GPT-4.1, Qwen 2.5 72B, and Llama 3.1 70B, respectively. Tables \ref{tab:ors_results_negative} and \ref{tab:ors_results_positive} show the feedback results in terms of $F_{0.5}$ for one random operational rater (OR), an ensemble of two random operational raters (OR2), and three random operational raters (OR3).

\begin{table}[ht!]

\centering
\begin{tabular}{lcc||c}
\hline
\textbf{Macro-aspect} & \textbf{Negative} & \textbf{Positive} & \textbf{All} \\
\hline
Language  & 45.71  & 47.14  & 77.86  \\
Content   & 48.57  & 49.29  & 79.29  \\
Attitude  & 27.14  & 31.43  & 54.29  \\
\hline
\hline
All       & 81.43  & 87.14  & 92.86  \\
\hline
\end{tabular}
\caption{Percentage of essays receiving feedback across macro-aspects.}
\label{tab:feed_percentage}
\end{table}

\begin{table*}[ht!]
\footnotesize
\centering
\begin{tabular}{lcccc|cccc}
\hline
 & \multicolumn{4}{c}{\textbf{Negative}} & \multicolumn{4}{c}{\textbf{Positive}} \\
\cline{2-9}
 & \textbf{P} & \textbf{R} & \textbf{\bm{$F_{0.5}$}} & \textbf{Prevalence}
 & \textbf{P} & \textbf{R} & \textbf{\bm{$F_{0.5}$}} & \textbf{Prevalence} \\
\hline
Int      & $47.54_{\pm 28.59}$  & $37.89_{\pm 9.05}$ & $40.77_{\pm 15.37}$  & 13.57  & $43.39_{\pm 15.12}$  & $29.17_{\pm 11.18}$  & $35.59_{\pm 5.34}$  & 17.14  \\
Cpl           & $34.25_{\pm 13.44}$  & $34.74_{\pm 13.56}$  & $30.76_{\pm 8.01}$ & 13.57  & $60.66_{\pm 23.40}$  & $23.64_{\pm 7.27}$  & $41.58_{\pm 6.13}$  & 15.71  \\
Acc             & $30.99_{\pm 14.06}$  & $31.11_{\pm 19.26}$  & $26.10_{\pm 3.80}$  & 19.29 & $68.73_{\pm 25.64}$  & $13.33_{\pm 4.86}$  & $34.75_{\pm 8.60}$  & 17.14  \\
Flu              & $20.42_{\pm 6.94}$  & $55.00_{\pm 18.71}$  & $22.37_{\pm 5.45}$  & 20.00  & $48.48_{\pm 26.92}$  & $20.77_{\pm 10.20}$  & $30.23_{\pm 4.04}$  & 18.57  \\
\hline
Cpr     &  $51.23_{\pm 26.36}$  & $25.45_{\pm 12.40}$  & $34.10_{\pm 7.78}$  & 15.71  & $33.10_{\pm 6.91}$  & $24.17_{\pm 5.53}$  & $29.75_{\pm 3.13}$  & 17.14  \\
Lgc            & $24.55_{\pm 9.37}$  & $40.00_{\pm 24.58}$  & $23.67_{\pm 7.67}$  & 15.0  & $62.56_{\pm 21.50}$  & $17.39_{\pm 13.19}$  & $33.06_{\pm 3.64}$  & 16.43  \\
Sph        & $41.78_{\pm 30.56}$ & $24.54_{\pm 26.44}$  & $23.83_{\pm 6.11}$  & 15.71  & $33.01_{\pm 9.76}$  & $34.00_{\pm 23.54}$  & $29.51_{\pm 6.83}$  & 14.29  \\
Prp        & $38.72_{\pm 8.42}$  & $22.86_{\pm 9.71}$  & $33.25_{\pm 9.22}$  & 15.00 & $64.44_{\pm 29.71}$  & $22.86_{\pm 10.17}$  & $38.23_{\pm 5.27}$  & 15.00  \\
\hline
Wil           & $46.37_{\pm 28.31}$  & $26.15_{\pm 11.77}$  & $32.79_{\pm 7.61}$  & 18.57  & $33.29_{\pm 13.41}$  & $42.61_{\pm 20.83}$  & $32.81_{\pm 11.11}$  &  16.43 \\
Inv           & $62.20_{\pm 9.49}$  & $24.61_{\pm 7.54}$  & $45.87_{\pm 4.38}$  & 9.29  & $63.79_{\pm 30.62}$  & $28.33_{\pm 14.04}$  & $40.24_{\pm 7.25}$  & 17.14  \\
\hline
\end{tabular}
\caption{Feedback results in terms of best Precision, Recall, and $F_{0.5}$ (OR).}
\label{tab:feedback_random}
\end{table*}

\begin{table*}[ht!]
\small
\centering
\begin{tabular}{lcccc|cccc}
\hline
 & \multicolumn{4}{c}{\textbf{Negative}} & \multicolumn{4}{c}{\textbf{Positive}} \\
\cline{2-9}
 & \textbf{P} & \textbf{R} & \textbf{\bm{$F_{0.5}$}} & \textbf{Prevalence}
 & \textbf{P} & \textbf{R} & \textbf{\bm{$F_{0.5}$}} & \textbf{Prevalence} \\
\hline
Intelligibility      & 54.11  & 26.32  & 44.68  & 13.57  & 18.51  & 45.83  & 21.02  & 17.14 \\
Complexity           & 73.89  & 21.05  & 49.20   & 13.57  & 100.00   & 36.36   & 74.07  & 15.71   \\
Accuracy              & 21.42  & 62.96 & 24.68   & 19.29  & 31.52  & 25.00  & 29.96   & 17.14  \\
Fluency               &  24.10  & 17.86  & 22.52   & 20.00   & 17.80  & 30.77   & 19.44   & 18.57  \\
\hline
Comprehensibility     & 41.70   & 27.27  & 37.71   & 15.71  & 20.59  & 58.33   & 23.65   & 17.14   \\
Logicality            & 100.00   & 14.29  & 45.45  & 15.00 & 18.44  & 52.17 & 21.18  & 16.43  \\
Sophistication        & 18.50 & 36.36  & 20.52  & 15.71  & 37.21  & 40.00  & 37.74  & 14.29  \\
Purposefulness        & 36.40 & 47.62  & 38.20  & 15.00 & 32.08  & 14.29  & 25.68  & 15.00  \\
\hline
Willingness           & 15.78  & 19.23  & 16.37  & 18.57 & 32.02  & 21.74  & 29.25  & 16.43  \\
Involvement           & 44.88  & 23.08  & 37.75  & 9.29  & 47.23  & 20.83  & 37.68  & 17.14  \\
\hline
\end{tabular}
\caption{Feedback results in terms of best Precision, Recall, and $F_{0.5}$ (GPT-4.1).}
\label{tab:feedback_gpt}
\end{table*}

\begin{table*}[ht!]
\small
\centering
\begin{tabular}{lcccc|cccc}
\hline
 & \multicolumn{4}{c}{\textbf{Negative}} & \multicolumn{4}{c}{\textbf{Positive}} \\
\cline{2-9}
 & \textbf{P} & \textbf{R} & \textbf{\bm{$F_{0.5}$}} & \textbf{Prevalence}
 & \textbf{P} & \textbf{R} & \textbf{\bm{$F_{0.5}$}} & \textbf{Prevalence} \\
\hline
Intelligibility      & 32.05 & 31.58  & 31.96  & 13.57  & 11.38 & 91.67  & 13.80  & 17.14 \\
Complexity           & 36.15  & 46.11  & 37.20  & 13.57  & 100.00  & 27.27  & 65.22  & 15.71 \\
Accuracy              & 13.88  & 96.30  & 16.75  & 19.29  & 34.94  & 29.17  & 33.61  & 17.14 \\
Fluency               & 15.09  & 57.14  & 17.70  & 20.00  & 17.02  & 30.77  & 18.69  & 18.57 \\
\hline
Comprehensibility     & 37.34  & 27.27  & 34.77  & 15.71  & 11.70  & 79.17  & 14.11  & 17.14 \\
Logicality            & 33.50  & 57.14  & 36.52  & 15.00  & 53.06  & 17.39  & 37.63  & 16.43 \\
Sophistication        & 37.34  & 13.64  & 27.71  & 15.71  & 66.67  & 15.00  & 39.47  & 14.29 \\
Purposefulness        & 71.58  & 19.05  & 46.13  & 15.00  & 28.24  & 23.81  & 27.23  & 15.00 \\
\hline
Willingness           & 10.47  & 92.31  & 12.72  & 18.57  & 36.11  & 17.39  & 29.71  & 16.43 \\
Involvement           & 68.46  & 15.38  & 40.51  & 9.29   & 36.94 & 50.00  & 38.98  & 17.14 \\
\hline
\end{tabular}
\caption{Feedback results in terms of best Precision, Recall, and $F_{0.5}$ (Qwen 2.5 72B).}
\label{tab:feedback_qwen}
\end{table*}

\begin{table*}[ht!]
\small
\centering
\begin{tabular}{lcccc|cccc}
\hline
 & \multicolumn{4}{c}{\textbf{Negative}} & \multicolumn{4}{c}{\textbf{Positive}} \\
\cline{2-9}
 & \textbf{P} & \textbf{R} & \textbf{\bm{$F_{0.5}$}} & \textbf{Prevalence}
 & \textbf{P} & \textbf{R} & \textbf{\bm{$F_{0.5}$}} & \textbf{Prevalence} \\
\hline
Intelligibility      & 67.98  & 15.79  & 40.92  & 13.57  & 19.96  & 54.17  & 22.84  & 17.14 \\
Complexity           & 34.67  & 31.58  & 34.00  & 13.57  & 100.00  & 31.82  & 70.00  & 15.71 \\
Accuracy              & 23.66  & 29.63  & 24.66  & 19.29  & 30.92  & 20.93  & 28.19  & 17.14 \\
Fluency               & 19.16  & 28.57  & 20.51  & 20.00  & 23.34  &  19.23 & 22.38  & 18.57 \\
\hline
Comprehensibility     & 28.43  & 18.18  & 25.55  & 15.71  & 20.28  & 75.00  & 23.74  & 17.14 \\
Logicality            & 26.46  & 38.10  & 28.18  & 15.00  & 44.17  & 30.43  & 40.52  & 16.43 \\
Sophistication        & 22.96  & 13.64  & 20.20  & 15.71  & 100.00  & 10.00  & 35.71  & 14.29 \\
Purposefulness        & 59.50  & 33.33  & 51.43  & 15.00  & 44.04  & 23.81  & 37.64  & 15.00 \\
\hline
Willingness           & 21.78  & 15.38  &  20.11 & 18.57  & 31.14  & 17.39  & 26.89  & 16.43 \\
Involvement           & 40.85  & 52.85  & 42.93  & 9.29   & 61.70  & 25.00  & 47.70  & 17.14 \\
\hline
\end{tabular}
\caption{Feedback results in terms of best Precision, Recall, and $F_{0.5}$ (Llama 3.1 70B).}
\label{tab:feedback_llama}
\end{table*}

\begin{table*}[ht!]
\footnotesize
\centering
\begin{tabular}{lccc|ccc}
\hline
 & \textbf{OR} & \textbf{OR2} & \textbf{OR3} & \textbf{GPT4.1} & \textbf{Qwen2.5} & \textbf{Llama3.1} \\
\hline
Int & $40.77_{\pm 15.37}$ & $34.06_{\pm 3.71}$  & $45.30_{\pm 12.82}$ & 44.68 & 31.96 & 40.92 \\
Cpl & $30.76_{\pm 8.01}$  & $41.77_{\pm 9.13}$  & $47.32_{\pm 6.78}$  & 49.20 & 37.20 & 34.00 \\
Acc & $26.10_{\pm 3.80}$  & $33.71_{\pm 3.47}$  & $44.75_{\pm 8.62}$  & 24.68 & 16.75 & 24.66 \\
Flu & $22.37_{\pm 5.45}$  & $41.12_{\pm 18.23}$ & $44.13_{\pm 6.46}$  & 22.52 & 17.70 & 20.51 \\
\hline
Cpr & $34.10_{\pm 7.78}$  & $38.42_{\pm 14.26}$ & $41.35_{\pm 9.32}$  & 37.71 & 15.71 & 25.55 \\
Lgc & $23.67_{\pm 7.67}$  & $32.01_{\pm 6.07}$  & $51.20_{\pm 5.14}$  & 45.45 & 36.52 & 28.18 \\
Sph & $23.83_{\pm 6.11}$  & $23.64_{\pm 5.44}$  & $40.02_{\pm 9.57}$  & 20.52 & 27.71 & 20.20 \\
Prp & $33.25_{\pm 9.22}$  & $44.81_{\pm 8.60}$  & $59.79_{\pm 9.10}$  & 38.20 & 46.13 & 51.43 \\
\hline
Wil & $32.79_{\pm 7.61}$ & $33.16_{\pm 4.57}$  & $38.74_{\pm 10.10}$ & 16.37 & 12.72 & 20.11 \\
Inv & $45.87_{\pm 4.38}$ & $49.22_{\pm 15.24}$ & $52.20_{\pm 10.90}$ & 37.75 & 40.51 & 42.93 \\
\hline
Avg. & 31.35 & 37.19 & 46.48 & 33.71 & 28.29 & 30.85 \\
\hline
\end{tabular}
\caption{Negative feedback results in terms of best $F_{0.5}$. OR, OR2, OR3 vs GPT4.1, Qwen2.5, Llama3.1.}
\label{tab:ors_results_negative}
\end{table*}

\begin{table*}[ht!]
\footnotesize
\centering
\begin{tabular}{lccc|ccc}
\hline
 & \textbf{OR} & \textbf{OR2} & \textbf{OR3} & \textbf{GPT4.1} & \textbf{Qwen2.5} & \textbf{Llama3.1} \\
\hline
Int & $35.59_{\pm 5.34}$ & $35.59_{\pm 5.60}$  & $39.92_{\pm 5.71}$ & 21.02 & 13.80 & 22.84 \\
Cpl & $41.58_{\pm 6.13}$ & $36.92_{\pm 5.66}$  & $56.44_{\pm 11.69}$ & 74.07 & 65.22 & 70.00 \\
Acc & $34.75_{\pm 8.60}$ & $45.15_{\pm 7.61}$  & $50.28_{\pm 8.58}$ & 29.96 & 33.61 & 28.19 \\
Flu & $30.23_{\pm 4.04}$ & $33.46_{\pm 7.41}$  & $42.79_{\pm 5.08}$ & 19.44 & 18.69 & 22.38 \\
\hline
Cpr & $29.75_{\pm 3.13}$ & $30.29_{\pm 6.46}$  & $45.63_{\pm 10.32}$ & 23.65 & 14.11 & 23.74 \\
Lgc & $33.06_{\pm 3.64}$ & $38.87_{\pm 9.27}$  & $48.76_{\pm 8.13}$ & 21.18 & 37.63 & 40.52 \\
Sph & $29.51_{\pm 6.83}$ & $28.29_{\pm 4.51}$  & $36.85_{\pm 9.96}$ & 37.74 & 39.47 & 35.71 \\
Prp & $38.23_{\pm 5.27}$ & $35.08_{\pm 5.54}$  & $45.71_{\pm 13.48}$ & 25.68 & 27.23 & 37.64 \\
\hline
Wil & $32.81_{\pm 11.11}$ & $50.18_{\pm 10.59}$ & $50.21_{\pm 12.03}$ & 29.25 & 29.71 & 26.89 \\
Inv & $40.24_{\pm 7.25}$  & $51.79_{\pm 6.22}$  & $61.79_{\pm 6.49}$ & 37.68  & 38.98 & 47.70 \\
\hline
Avg. & 34.57 & 38.56 & 47.84 & 31.96  & 31.84 & 35.56 \\
\hline
\end{tabular}
\caption{Positive feedback results in terms of best $F_{0.5}$. OR, OR2, OR3 vs GPT4.1, Qwen2.5, Llama3.1.}
\label{tab:ors_results_positive}
\end{table*}

\end{document}